\documentclass[10pt,twocolumn,letterpaper]{article}

\usepackage{iccv}
\usepackage{times}
\usepackage{epsfig}
\usepackage{graphicx}
\usepackage{amsmath}
\usepackage{amssymb}
\usepackage{subfigure}
\usepackage{comment}
\usepackage{color}
\usepackage{enumitem}
\usepackage{multirow}
\usepackage{bbm}
\usepackage[pagebackref=true,breaklinks=true,letterpaper=true,colorlinks,bookmarks=false]{hyperref}

 \iccvfinalcopy 

\def\iccvPaperID{7307}

\ificcvfinal\fi

\begin{document}

\title{Video-based Person Re-identification with Spatial and Temporal Memory Networks}

\author{
Chanho Eom \quad\quad\quad Geon Lee \quad\quad\quad Junghyup Lee \quad\quad\quad Bumsub Ham\thanks{Corresponding author.}\vspace*{0.2cm}\\
{School of Electrical and Electronic Engineering, Yonsei University}\\
\url{https://cvlab-yonsei.github.io/projects/STMN}
}

\maketitle
\ificcvfinal\thispagestyle{empty}\fi

\begin{abstract}
\vspace{-0.3cm}
Video-based person re-identification~(reID) aims to retrieve person videos with the same identity as a query person across multiple cameras. Spatial and temporal distractors in person videos, such as background clutter and partial occlusions over frames, respectively, make this task much more challenging than image-based person reID. We observe that spatial distractors appear consistently in a particular location, and temporal distractors show several patterns,~e.g.,~partial occlusions occur in the first few frames, where such patterns provide informative cues for predicting which frames to focus on~(i.e.,~temporal attentions). Based on this, we introduce a novel Spatial and Temporal Memory Networks (STMN). The spatial memory stores features for spatial distractors that frequently emerge across video frames, while the temporal memory saves attentions which are optimized for typical temporal patterns in person videos. We leverage the spatial and temporal memories to refine frame-level person representations and to aggregate the refined frame-level features into a sequence-level person representation, respectively, effectively handling spatial and temporal distractors in person videos. We also introduce a memory spread loss preventing our model from addressing particular items only in the memories. Experimental results on standard benchmarks, including MARS, DukeMTMC-VideoReID, and LS-VID, demonstrate the effectiveness of our method.
	\vspace{-0.4cm}
\end{abstract}
		
		\begin{figure}
			\centering
			\includegraphics[width=0.95\linewidth]{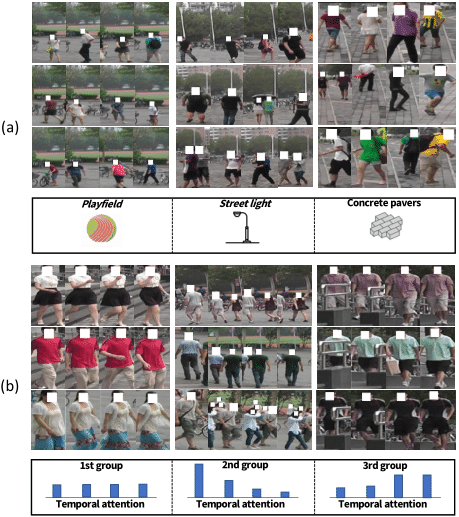}
			\caption{Examples of (a) spatial distractors that appear frequently in surveillance videos and (b) prototypes of temporal patterns that provide important clues to predict temporal attentions.}
			\label{fig:teaser}
			\vspace{-0.4cm}
		\end{figure}

\section{Introduction}
\vspace{-0.2cm}

Person re-Identification (reID) aims at retrieving a person of interest from a set of pedestrian images/videos taken from non-overlapping cameras. Convolutional neural networks~(CNNs) have made remarkable advances in image-based person reID~\cite{zhao2017spindle,su2017pose,liu2017hydraplus,li2018harmonious,eom2019learning,zheng2019joint} over the last decade. Video-based person reID has recently attracted increasing attention in accordance with the prevalence of video capturing systems. Video frames provide rich information to specify a particular person, but they often contain spatial distractors,~\eg,~trees, bicycles, and concrete pavers. In particular, person videos, typically cropped by off-the-shelf object detectors from a whole sequence, also have temporal distractors,~\eg,~misaligned persons across video frames or partial occlusions within a sequence.

Recent video reID methods~\cite{li2018diversity,fu2019sta} attempt to tackle these issues by exploiting spatial and temporal attention modules, which are useful for extracting person representations robust to noisy regions~(\eg, background clutter) and temporal variations~(\eg, partial occlusions). They, however, do not consider a global view in a sequence~\cite{zhang2020multi,hou2020temporal}, suggesting that these approaches may focus on less discriminative parts or video frames. Several works~\cite{li2019multi,li2019global,liu2019spatially,subramaniam2019co,yan2020learning,yang2020spatial} instead propose to use non-local~\cite{wang2018non} or graph convolutional networks~\cite{kipf2017semi} to capture co-attention over frames. They focus on the shared information across multiple frames to obtain a person representation from a video, taking into account the temporal context. The co-attention, however, may concentrate on distracting scene details or partial occlusions, which are often shared in successive video frames, producing an incorrect video representation.

We present in this paper Spatial and Temporal Memory Networks~(STMN) to extract person representations robust to spatial and temporal distractors for video-based person reID. The main idea is based on the following observations:~1) Since video sequences are captured by a stationary camera, it is likely that they constantly contain background clutter such as a playfield, a street light, or concrete pavers in a particular location (Fig.~\ref{fig:teaser}(a));~2) Temporal patterns,~\eg,~a person of interest disappears at the end of the sequence~(Fig.~\ref{fig:teaser}(b) center) or partial occlusions occur in the first few frames (Fig.~\ref{fig:teaser}(b) right), provide crucial clues to determine which frames we have to focus on~(\ie,~temporal attentions).

Based on the observations, we propose to exploit two external memories called spatial and temporal memories. The spatial memory is trained to store spatial distractors that frequently appear across video frames, while the temporal memory is trained to memorize attentions which are optimized for typical temporal patterns in person videos. At test time, we leverage the memories as look-up tables, and ease the difficulty of handling the spatial and temporal distractors from videos of unseen identities. Specifically, we exploit the spatial memory to suppress features for distracting scene details from each frame-level person representation, and the temporal one to aggregate the frame-level person representations focusing more on discriminative frames. We also propose a memory spread loss that encourages our model to access all items in the memories during training. We demonstrate the effectiveness of our method on the MARS~\cite{zheng2016mars}, DukeMTMC-VideoReID~\cite{wu2018exploit}, and LS-VID~\cite{li2019global} datasets. To our best knowledge, this is an early effort that jointly leverages multiple types of memories. The main contributions of our work can be summarized as follows:

	\vspace{-0.1cm}	
	\begin{itemize}[leftmargin=*]
		\item[$\bullet$] We introduce a simple yet effective method for video-based person reID, dubbed STMN, which extracts a robust video representation to spatial and temporal distractors using spatial and temporal memories.
		\vspace{-0.1cm}
		\item[$\bullet$] We propose a memory spread loss that prevents our model from accessing few items repeatedly, encouraging all items in the memories to be used.
		\vspace{-0.1cm}
		\item[$\bullet$] We achieve the state of the art on standard video reID benchmarks.  Ablation studies further validate the effectiveness of our method.
	\end{itemize}
	

\section{Related Work}
\vspace{-0.2cm}
	
	Here, we briefly introduce representative works closely related to ours, and clarify their differences from ours.
	
	\vspace{-0.4cm}
	\paragraph{Video-based person reID.}
	
		The key for video-based reID is to extract person representations robust to spatial and temporal distractors. Many methods~\cite{liu2017quality,li2018diversity,fu2019sta} propose to use attention modules for video-based person reID. QAN~\cite{liu2017quality} uses a temporal attention to aggregate frame-level features, focusing on discriminative frames. DRSA~\cite{li2018diversity} and STA~\cite{fu2019sta} additionally use a spatial attention to suppress features for spatial distractors. They, however, assign attentions to each frame without considering whole frames in a sequence, indicating that they may aggregate less discriminative parts or frames in the sequence~\cite{zhang2020multi,hou2020temporal}. Recent methods~\cite{li2019multi,li2019global,liu2019spatially,subramaniam2019co,yan2020learning} propose to use co-attention modules between frames by adopting non-local~\cite{wang2018non} or graph convolutional networks~\cite{kipf2017semi}. Specifically, GLTR~\cite{li2019global} adds a co-attention module at the end of backbone CNNs, while M3D~\cite{li2019multi}, STE-NVAN~\cite{liu2019spatially} and COSAM~\cite{subramaniam2019co} insert multiple co-attention modules into different levels of backbone CNNs, to refine frame-level person representations, considering contextual temporal relations between frames. The work of \cite{yan2020learning,yang2020spatial} introduces a hierarchical co-attention module, dividing frames into multiple granularities, to capture discriminative spatial and temporal features from different semantic levels. These approaches highlight shared information between frames, suppressing features from distracting scene details and occlusion, which is useful only when such distractors appear in a few frames. When similar backgrounds and/or occlusion are shared across frames, the features from these distractors are propagated, which rather interferes with retrieving persons. The works of~\cite{mclaughlin2016recurrent,yan2016person,zhou2017see} propose to use recurrent neural networks (RNNs) for aggregating frame-level person representations robust to temporal distractors. The hidden states of RNNs store the temporal context in previous frames, and allow to aggregate the person representations selectively, based on the context. We also exploit RNNs in STMN, but we do not use them directly to aggregate frame-level representations, which may be suboptimal, since RNNs do not consider a temporal context in whole frames (except at the last time step). We instead leverage RNNs to encode a temporal pattern of a sequence for accessing a temporal memory.

		Previous works overlook the fact that several scene details and temporal patterns repeatably appear in surveillance videos, which may provide important cues to handle spatial and temporal distractors. STMN stores the scene details and attentions for the temporal patterns in the spatial and temporal memories, respectively, providing person representations robust against the spatial and temporal distractors.
	
	\vspace{-0.4cm}
	\paragraph{Memory network.}
	
		The work of~\cite{weston2014memory} first introduces memory networks to handle long-term dependencies for question and answering. They, however, require extra supervisory signals to access the memory, and are not able to be trained end-to-end. The soft addressing technique~\cite{sukhbaatar2015end} addresses these problems by using attention maps to access the memory. Key-value memory networks~\cite{miller2016key} propose to adopt different encodings for accessing and reading operations, where they address relevant memory items by keys, and their corresponding values are subsequently returned. Recently, many computer vision methods exploit memory networks for, \eg, one-shot learning~\cite{cai2018memory}, video object segmentation~\cite{oh2019video}, domain adaptation~\cite{zhong2019invariance}, image colorization~\cite{yoo2019coloring}, and anomaly detection~\cite{gong2019memorizing,park2020learning}. Our work also leverages memory networks but for recording features for distracting scene details and temporal attentions. By using the memory networks, we are able to extract person representations robust against spatial and temporal distractors. In addition, we propose a memory spread loss to penalize our model when it keeps accessing particular items only, while other items remain unused.

		
\vspace{-0.3cm}
\section{Approach}
\vspace{-0.2cm}

	In this section, we provide a brief overview of our approach to exploiting spatial and temporal memories for video-based reID (Sec.~\ref{subsec:overview}). We then present a detailed description for a network architecture (Sec.~\ref{subsec:network_architecture}) and training losses~(Sec.~\ref{subsec:training_loss}).

	\vspace{-0.1cm}
	\subsection{Overview} \label{subsec:overview}
	\vspace{-0.2cm}
	
\begin{figure}
			\centering
			\includegraphics[height=0.17\textheight,width=0.95\linewidth]{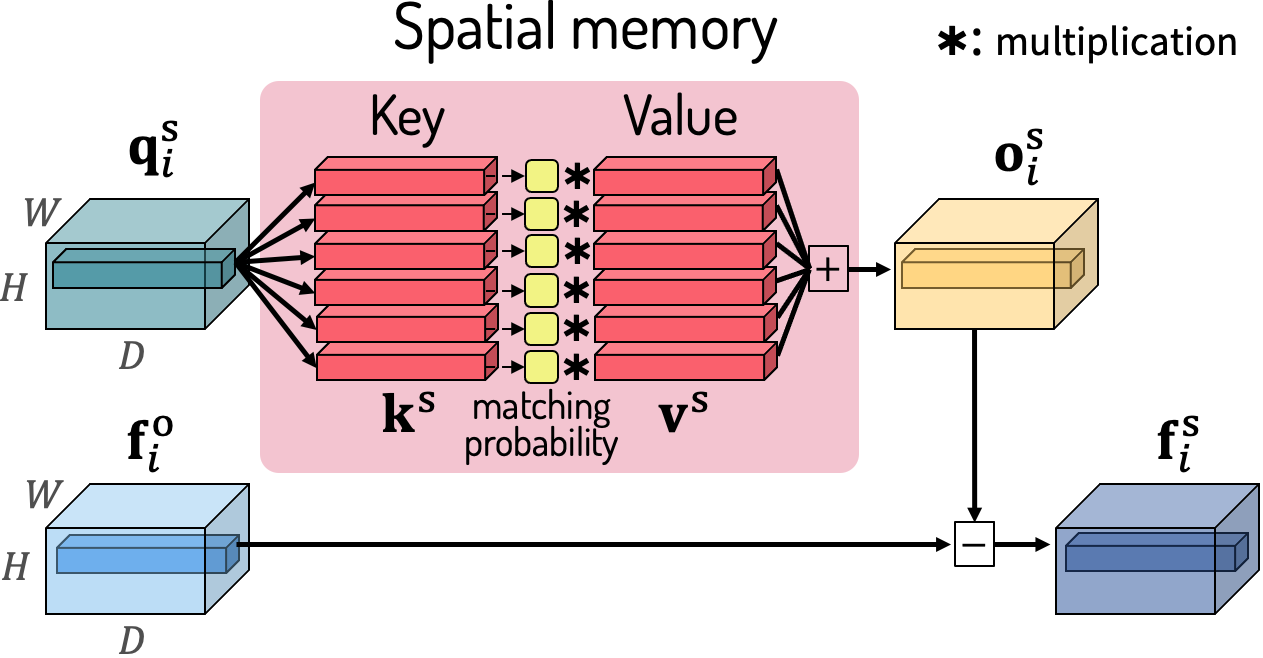}
		\caption{The spatial memory takes a person representation $\mathbf{f}^{\text o}_i \in \mathbb{R}^{D \times K}$ and a query map $\mathbf{q}^{\text s}_i \in \mathbb{R}^{D \times K}$ of the $i$-th frame as inputs. We access the memory based on the matching probability between the query feature $\mathbf{q}^{\text s}_{i,k} \in \mathbb{R}^{D}$ and keys $\mathbf{k}^{\text s}$, and use the output to refine the input representation $\mathbf{f}^{\text o}_{i,k} \in \mathbb{R}^{D}$. (Best viewed in color.)}
		\vspace{-0.5cm}
		\label{fig:spatial_memory}
		\end{figure}

\begin{figure*}
			\centering
			\includegraphics[height=0.17\textheight,width=0.95\linewidth]{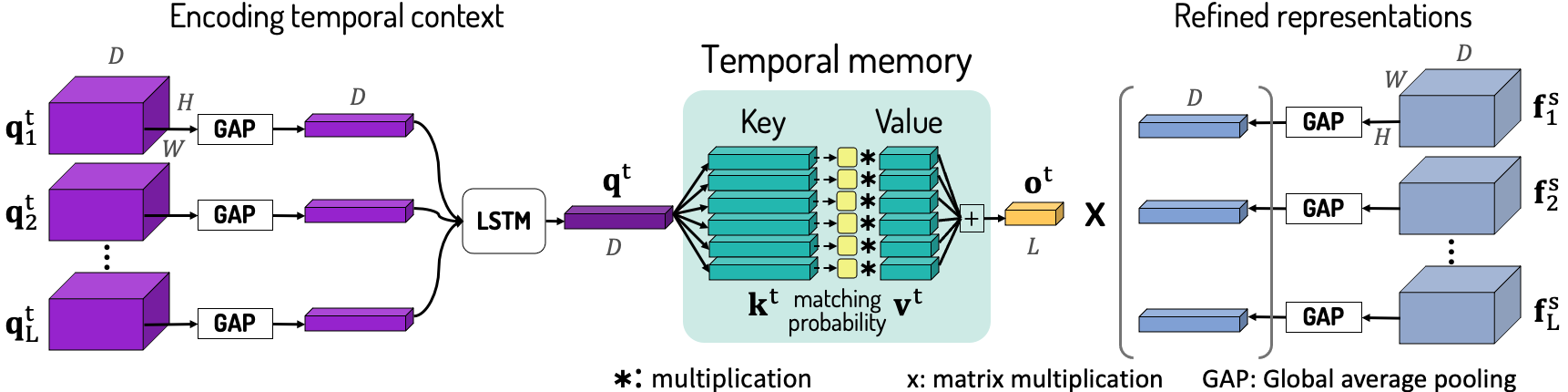}
		\caption{The temporal memory takes a sequence of query maps $\mathbf{q}^\text{t}_i |_{i=1}^L$ and the person representations $\mathbf{f}^\text{s}_i |_{i=1}^L$ that are refined by the spatial memory as inputs. We aggregate the query maps by using global average pooling and LSTM modules, and use the output to address the memory. The memory outputs temporal attentions $\mathbf{o}^{\text t}$, and the attentions are used to aggregate the frame-level representations into a sequence-level one. (Best viewed in color.)}
		\vspace{-0.5cm}
		\label{fig:temporal_memory}
		\end{figure*}

		STMN mainly consists of three components: an encoder, a spatial memory~(Fig.~\ref{fig:spatial_memory}), and a temporal memory~(Fig.~\ref{fig:temporal_memory}). For each frame, the encoder extracts a person representation and two query maps, where each query is used to access either spatial or temporal memories. The spatial memory stores features for scene details, frequently appearing across video frames, such as street lights, trees, and concrete pavers. We extract such features from the spatial memory using the corresponding query map, and use them to refine the person representation, removing information that interferes with identifying persons. The temporal memory saves attentions optimized for typical temporal patterns that repeatably occur in person videos. We access the temporal memory with the corresponding query map, and use the output to aggregate the refined frame-level features into a sequence-level person representation. We train our model end-to-end using memory spread, triplet, and cross-entropy terms.
		
	\vspace{-0.2cm}
	\subsection{Network architecture} \label{subsec:network_architecture}
	
		\vspace{-0.1cm}
		\paragraph{Encoder.} \label{subsubsec:encoder}
		
			The encoder takes a video sequence $F_i|_{i=1}^L$ as an input, where $F_i$ is the $i$-th frame of the sequence, and $L$ is the total number of frames. We exploit ResNet~\cite{he2016deep} cropped at \texttt{conv4} layer as our backbone network, where the network parameters are pre-trained for ImageNet classification~\cite{krizhevsky2012imagenet}. We add three heads on top of the backbone network to extract feature maps for each frame: a frame-level person representation $\mathbf{f}^{\text o}_i$, and query maps, $\mathbf{q}^{\text s}_i$ and $\mathbf{q}^{\text t}_i$, for accessing the spatial and temporal memories, respectively. Each feature map has a size of $D \times H \times W$, where $D$, $H$, and $W$ are the number of channels, height, and width, respectively. We denote by $\mathbf{f}^{\text o}_{i,k}$, $\mathbf{q}^{\text s}_{i,k}$, and $\mathbf{q}^{\text t}_{i,k}$ individual features of size $D$ at position $k$, where $k \in \{ 1, 2, ..., K \}$ and $K = H \times W$.

		\vspace{-0.4cm}
		\paragraph{Spatial memory.} \label{subsubsec:spatial_memory}
		
			Frame-level person representations extracted by the encoder may contain features for distracting scene details~(\emph{e.g.},~trees, concrete paver, bicycles, or cars), which may prevent distinguishing different pedestrians in similar scenes. To handle this problem, we refine the frame-level person representations  using a spatial memory~(Fig.~\ref{fig:spatial_memory}).
			
			The spatial memory has a key-value structure, and contains $M$ items. The values  $\mathbf{v}^{\text s} \in \mathbb{R}^{D \times M}$ encode distracting scene details over the video sequence, while the keys $\mathbf{k}^{\text s} \in \mathbb{R}^{D \times M}$ are used to access corresponding values. We denote by $\mathbf{k}^{\text s}_n \in \mathbb{R}^{D}$ and $\mathbf{v}^{\text s}_n \in \mathbb{R}^{D}$ each key and value in the memory, respectively, where $n \in \{ 1, 2, ... , M\} $. The spatial memory takes a person representation $\mathbf{f}^{\text o}_i \in \mathbb{R}^{D \times K}$ and a query map $\mathbf{q}^{\text s}_i \in \mathbb{R}^{D \times K}$ of the frame $F_i$ as inputs. Since different parts of the input frame may contain distinct scene details, we access the memory with individual components of the input query map, $\mathbf{q}^{\text s}_{i,k} \in \mathbb{R}^{D}$. Specifically, we compute cosine similarities between the query $\mathbf{q}^{\text s}_{i,k}$ and all keys $\mathbf{k}^{\text s}$ in the memory, resulting in a correlation map of size $1 \times M$. We then normalize it as follows:
					\begin{equation}
						a^{\text s}_{i,k,n} = \frac{\exp((\mathbf{q}^{\text s}_{i,k})^{\text T}\ \mathbf{k}^{\text s}_n)}{\sum_{n^\prime=1}^{M} \exp((\mathbf{q}^{\text s}_{i,k})^{\text T}\ \mathbf{k}^{\text s}_{n^\prime})}.
					\label{eq:a_s}
					\end{equation}
			The matching probability $a^{\text s}_{i,k,n}$ represents a likelihood that the scene detail recorded in the $n$-th memory item exists in the $k$-th position of the $i$-th frame. The memory outputs a weighted average of values $\mathbf{v}^{\text s}_n$ using the corresponding probabilities $a^{\text s}_{i,k,n}$ as follows:
					\begin{equation}
						\mathbf{o}^{\text s}_{i,k} = \sum_{n=1}^{M} a^{\text s}_{i,k,n} \mathbf{v}^{\text s}_n,
					\end{equation}
			where the output of the spatial memory, $\mathbf{o}^{\text s}_{i,k}$, contains uninformative features, that interfere with identifying persons, for the $k$-th position of the $i$-th frame. We use the output of the spatial memory to refine the person representation as follows:
					\begin{equation}
						\mathbf{f}^{\text s}_{i,k} = \mathbf{f}^{\text o}_{i,k} - \text {BN}(\mathbf{o}^{\text s}_{i,k}).
						\label{eq:refine}
					\end{equation}
			Motivated by \cite{wang2018non}, we use a batch normalization (BN) layer to adjust the distribution gap between outputs from the encoder and the spatial memory. Note that our spatial memory is similar to non-local networks~\cite{wang2018non} in that they both refine input features in a residual manner. However, ours is clearly different from the non-local networks. Keys and values in our method are external parameters stored in the memory, and they are updated by backpropagation during training in order to memorize the scene details. On the contrary, keys, queries, and values in the non-local networks are computed from input features, similar to a self-attention method~\cite{vaswani2017attention}. 		
			
		\vspace{-0.4cm}
		\paragraph{Temporal memory.} \label{subsubsec:temporal_memory}
		
			The refinement process using the spatial memory operates on each frame independently, which is not capable of capturing temporal contexts in video sequences. This may lead our framework susceptible to occlusion or misalignment between frames. To address this problem, we propose to use an additional temporal memory network~(Fig.~\ref{fig:temporal_memory}).
			
			The temporal memory also has a key-value structure, and contains $N$ items, where the keys $\mathbf{k}^{\text t} \in \mathbb{R}^{D \times N}$ encode prototypes of temporal patterns that repeatably appear in person videos, and the values $\mathbf{v}^{\text t} \in \mathbb{R}^{L \times N}$ memorize temporal attentions which are optimized for the corresponding temporal patterns. We denote by $\mathbf{k}^{\text t}_n \in \mathbb{R}^{D}$ and $\mathbf{v}^{\text t}_n \in \mathbb{R}^{L}$ each key and value in the memory, respectively, where $n \in \{ 1, 2, ... , N\} $. The temporal memory takes a sequence of query maps $\mathbf{q}^\text{t}_i |_{i=1}^L$ and the person representations refined by the spatial memory $\mathbf{f}^\text{s}_i |_{i=1}^L$ as inputs. We first encode a temporal context of a given sequence,~\emph{e.g.},~the occlusion arises in the middle frame, using the query maps. Concretely, we spatially aggregate the input query maps by a global average pooling (GAP), and feed them into a long short-term memory (LSTM)~\cite{hochreiter1997long} as follows:
					\begin{equation}
						\mathbf{q}^{\text t} = \text{LSTM}([\text{GAP}(\mathbf{q}^{\text t}_{1}), \text{GAP}(\mathbf{q}^{\text t}_{2}), ..., \text{GAP}(\mathbf{q}^{\text t}_{L})]),
					\end{equation}
			where $\mathbf{q}^{\text t} \in \mathbb{R}^{D}$ is an output of the last time step, representing the temporal context of the sequence. We then use the temporal context $\mathbf{q}^{\text t}$ to access the temporal memory in a similar way to the spatial one as follows:
					\begin{equation}
						a^{\text t}_{n} = \frac{\exp((\mathbf{q}^{\text t})^{\text T}\ \mathbf{k}^{\text t}_n)}{\sum_{n^\prime=1}^{N} \exp((\mathbf{q}^{\text t})^{\text T}\ \mathbf{k}^{\text t}_{n^\prime})},
					\label{eq:a_t}
					\end{equation}
			where $a^{\text t}_{n}$ represents a probability that the encoded temporal context $\mathbf{q}^{\text t}$ belongs to the temporal pattern stored in the $n$-th memory item~$\mathbf{k}^{\text t}_n$. We synthesize a temporal attention specific for the given sequence by taking weighted average over the values with the corresponding probability $a^{\text t}_{n}$ as follows:
					\begin{equation}
						\mathbf{o}^{\text t} = \sum_{n=1}^{N} a^{\text t}_{n} \mathbf{v}^{\text t}_n,
					\label{eq:att}
					\end{equation}
			where the memory output $\mathbf{o}^{\text t} \in \mathbb{R}^{L}$ represents the temporal attention, and $o^{\text t}_{i}$, the $i$-th element of the output, indicates the relative importance of the $i$-th frame in the sequence. We then apply a softmax function on the temporal attention $\mathbf{o}^{\text t}$, and use it to aggregate the refined frame-level features~$\mathbf{f}^{\text s}_i$ as follows:
					\begin{equation}
						\mathbf{f}^{\text t} = \sum_{i=1}^{L} \hat{o}^{\text t}_{i} \ \text{GAP}(\mathbf{f}^{\text s}_i),
					\end{equation}
			where $\hat{o}^{\text t}_{i} = \exp(o^{\text t}_{i})/\sum_{i^\prime=1}^{L}\exp(o^{\text t}_{i^\prime})$, and $\mathbf{f}^{\text t}$ is our final person representation for the input video sequence~$F_i|_{i=1}^L$.
			
			Note that previous methods,~\eg,~\cite{zhou2017see,li2018diversity,fu2019sta,li2019multi,li2019global,liu2019spatially,subramaniam2019co,yan2020learning}, decide which frames to focus on during temporal fusion based on person representations. This may enforce the representations to encode temporal contexts as well as identity-related cues, preventing the representations from being discriminative, particularly when video sequences of different identities contain similar temporal contexts. In our framework, on the contrary, person representations are decoupled from encoding temporal contexts, where query maps $\mathbf{q}^\text{t}_i$ and keys in the temporal memory, $\mathbf{k}^{\text t}$, encode such contexts. This encourages our model to extract person representations focusing on information that is useful for discriminating different identities, leading to performance gains on the reID task.

	\vspace{-0.1cm}
	\subsection{Training loss} \label{subsec:training_loss}
	\vspace{-0.1cm}
	
		\begin{figure}
			\centering
			\includegraphics[height=0.22\textheight,width=0.95\linewidth]{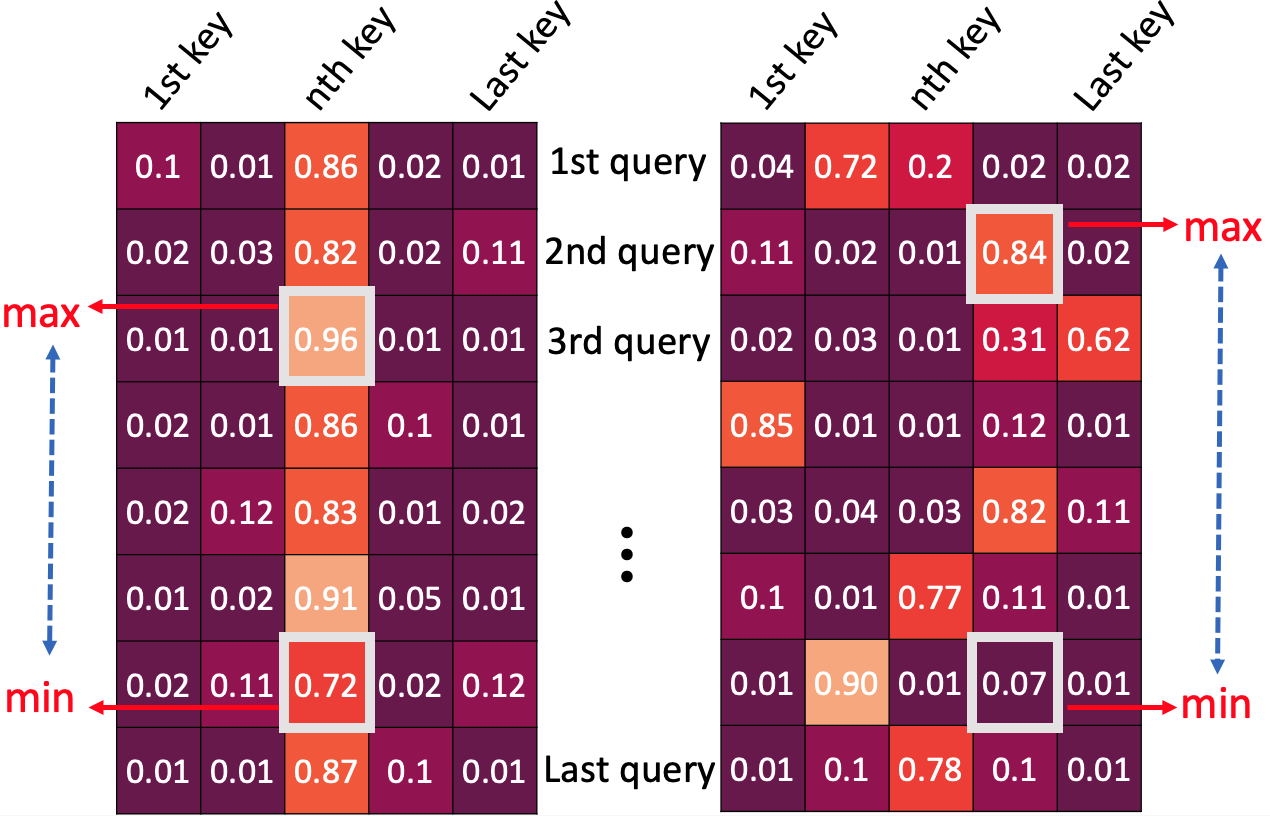}
			\caption{Example of matching probability maps for the case when our model addresses a particular memory item only (left) and the case when it uses all items in the memory (right). (Best viewed in color.)}
			\label{fig:mem_spread_loss}
			\vspace{-0.6cm}
		\end{figure}
	
		We use two terms to train our model end-to-end as follows:
			\begin{equation}
				\mathcal{L}_{total}	= \mathcal{L}_{S} + \mathcal{L}_{ID},
			\end{equation}
		where we denote by $\mathcal{L}_{S}$ and $\mathcal{L}_{ID}$ memory spread and identification losses, respectively. The memory spread term penalizes our model when it accesses a particular memory item only, while the identification term allows to extract discriminative person representations from video sequences. The detailed descriptions of each loss are presented in the following.
		
		\vspace{-0.4cm}
		\paragraph{Memory spread term.}
		
			We denote by $\mathbf{A}^{\text s} \in \mathbb{R}^{LKB \times M}$ and $\mathbf{A}^{\text t} \in \mathbb{R}^{B \times N}$ matching probability maps for the spatial and temporal memories, respectively, in a mini-batch, where $B$ is the number of sequences in the mini-batch. Note that we address the spatial and temporal memories $LKB$ and $B$ times for each mini-batch, respectively. Since we do not have extra supervisory signals except identification labels, we do not know which key should be matched to the input query. In this context, our model may address particular keys continually, while others are left unused~(Fig.~\ref{fig:mem_spread_loss} left). This causes memories to produce similar outputs regardless of input frames or sequences. To address this problem, we propose a memory spread loss as follows:
					\begin{equation} \label{eq:l_s}
						\begin{split}
							\mathcal{L}_{S} = \sum_{n=1}^{M} 
								[~ min (\mathbf{a}^{\text s}_n) 
								 - max (\mathbf{a}^{\text s}_n) + \alpha~]_+& \\
								+ [~ min (\mathbf{a}^{\text t}_n) 
								 - max (\mathbf{a}^{\text t}_n)& + \alpha~]_+,
						\end{split}
					\end{equation}
			where $\mathbf{a}^{\text s}_n \in \mathbb{R}^{LKB}$ and $\mathbf{a}^{\text t}_n \in \mathbb{R}^{B}$ are the $n$-th column vector of $\mathbf{A}^{\text s}$ and $\mathbf{A}^{\text t}$, respectively, representing matching probabilities of the $n$-th key in each memory w.r.t all queries in a mini-batch. $min(\cdot)$ and $max(\cdot)$ return the minimum and maximum values of an input vector. The memory spread loss enforces the minimum and maximum values of $\mathbf{a}^{\text s}_n$ and $\mathbf{a}^{\text t}_n$ to differ by at least a pre-defined margin $\alpha$. This prevents the case when our model keeps addressing a particular memory item~(Fig.~\ref{fig:mem_spread_loss} left), while encouraging it to access all memory items during training~(Fig.~\ref{fig:mem_spread_loss} right).
		
		\vspace{-0.4cm}
		\paragraph{Identification term.}
		
			Following other person reID methods~\cite{yan2020learning,zhang2020multi,hou2020temporal,chentemporal}, we exploit a combination of cross-entropy and batch-hard triplet~\cite{hermans2017defense} terms, with identification labels as a supervisory signal. The former encourages our model to learn a person representation $\mathbf{f}^{\text t}$ by focusing on identity-related cues, while the latter enforces the representations of the same identity to be closer to each other than those of different identities in the embedding space. Motivated by a deep supervision technique~\cite{lee2015deeply,wang2015training}, we also use the frame-level representations $\mathbf{f}^{\text s}_i |_{i=1}^L$ to compute the cross-entropy and triplet losses, where global and temporal average pooling are used to aggregate frame-level representations into a sequence-level one.
			
\vspace{-0.2cm}
\section{Experiments}
\vspace{-0.1cm}
	In this section, we provide implementation details of STMN (Sec.~\ref{subsec:implement}), and show ablation studies and visual analysis on spatial and temporal memories to validate the effectiveness of STMN (Sec.~\ref{subsec:discussion}). Lastly, we compare our method with the state of the art (Sec.~\ref{subsec:comp_w_sota}).

	\vspace{-0.1cm}
	\subsection{Implementation details}  \label{subsec:implement}
	
		\vspace{-0.2cm}
		\paragraph{Dataset and evaluation metric.}
		
			We evaluate our model on MARS~\cite{zheng2016mars}, DukeMTMC-VideoReID~\cite{ristani2016performance,wu2018exploit} (abbreviated as ``DukeV''), and LS-VID~\cite{li2019global}, following the standard protocol of each dataset. Note that we do not use PRID~\cite{hirzer2011person} and iLIDS-VID~\cite{wang2014person} for evaluation, since they contain few sequences captured with two cameras only. We report cumulative matching characteristics at rank-1 and mean average precision (mAP) for quantitative comparisons. 
		
		\vspace{-0.4cm}
		\paragraph{Training.}
		
			We train our model end-to-end for $200$ epochs using the Adam~\cite{kingma2014adam} optimizer, where $\beta_1$ and $\beta_2$ are set to $0.9$ and $0.999$, respectively. The learning rate, initially set to 1e-4, is reduced by a factor of 10 for every $50$ epochs. To train our model, we randomly choose $8$ identities, and sample $4$ sequences for each identity. Following the restricted random sampling (RRS) strategy~\cite{li2018diversity}, we then divide each sequence into $L$ chunks, and randomly choose one frame from each chunk. We resize input frames into the size of $256 \times 128$, and augment them with horizontal flipping and random erasing~\cite{zhong2017random}.

		\vspace{-0.4cm}		
		\paragraph{Hyperparameter.}
		
			To set the sizes of spatial and temporal memories, $M$ and $N$, the pre-defined margin $\alpha$ in the memory spread loss, and the length of an input sequence $L$, we divide the training set of MARS~\cite{zheng2016mars} into two subsets. Specifically, we randomly divide identities in the training set into two subsets of sizes $500/125$, and use corresponding $7075/1223$ sequences as training/validation splits. For query sequences, we randomly select $200$ sequences from the validation split. For the sizes of memories, we perform a grid search over $(M,N)$ pairs, where $M,N \in \{5, 10, 20 \}$. We choose a pair of $M=10$ and $N=5$ for our final model, which shows the best result in terms of the mean and standard deviation of rank-1 accuracy and mAP for five trials. For the margin $\alpha$ and the sequence length $L$, we also use a grid search over $\alpha \in \{ 0.1, 0.3, 0.5, 0.7, 1.0\}$ and $L \in \{ 4, 6, 8, 10\}$, respectively, setting $\alpha = 0.3$ and $L = 6$. We fix all hyperparameters, and train our model on the training splits of MARS~\cite{zheng2016mars}, DukeV~\cite{wu2018exploit} and LS-VID~\cite{li2019global}. Please refer to the supplementary materials for the details.
		
		\setlength{\tabcolsep}{4pt}
		\begin{table}[!t]
			\centering
				\resizebox{0.9\linewidth}{!}{
					\begin{tabular}{lccccccc}
					\hline\noalign{\smallskip}
					\multicolumn{1}{c}{\multirow{2}*{Methods}} & \multicolumn{2}{c}{MARS} & \multicolumn{2}{c}{DukeV} & \multicolumn{2}{c}{LS-VID} \\
					 & R-1 & mAP & R-1 & mAP & R-1 & mAP \\
					\noalign{\smallskip}
					\hline
					\noalign{\smallskip}
					\textcircled{\raisebox{-0.9pt}{1}} Baseline  							& 87.3 & 79.1 & 95.0 & 92.7 & 71.6 & 55.9 \\
					\textcircled{\raisebox{-0.9pt}{2}}~~~+ SM (w/o $\mathcal{L}_{S}$)		& 88.7 & 81.6 & 95.4 & 93.6 & 78.8 & 64.7 \\
					\textcircled{\raisebox{-0.9pt}{3}}~~~+ SM 								& 89.3 & \underline{82.5} & \underline{96.2} & \underline{94.2} & 79.6 & \underline{65.8} \\
					\textcircled{\raisebox{-0.9pt}{4}}~~~+ TM  (w/o $\mathcal{L}_{S}$)		& 88.5 & 81.9 & 95.2 & 93.3 & 77.8 & 63.0 \\
					\textcircled{\raisebox{-0.9pt}{5}}~~~+ TM								& \underline{89.5} & 82.0 & 95.4 & 93.7 & 78.9 & 64.4 \\
					\textcircled{\raisebox{-0.9pt}{6}}~~~+ SM + TM (w/o $\mathcal{L}_{S}$)	& 89.1 & 81.9 & 95.6 & 93.9 & \underline{79.9} & 65.4 \\
					\textcircled{\raisebox{-0.9pt}{7}}~~~+ SM + TM 						& \textbf{89.9} & \textbf{83.7} & \textbf{96.7} & \textbf{94.6} & \textbf{80.6} & \textbf{66.6} \\
					\hline
					\end{tabular}
				}
			\vspace{+0.1cm}
			\caption{Quantitative comparison for variants of our model on MARS~\cite{zheng2016mars}, DukeV~\cite{wu2018exploit} and LS-VID~\cite{li2019global}. Numbers in bold indicate the best performance and underscored ones are the second best. SM: spatial memory; TM: temporal memory.}
			\label{table:ablation_studies}
		\end{table}
		\setlength{\tabcolsep}{1.4pt}
				
		\begin{figure}
			\vspace{-0.2cm}
			\centering
			\includegraphics[height=0.15\textheight,width=\linewidth]{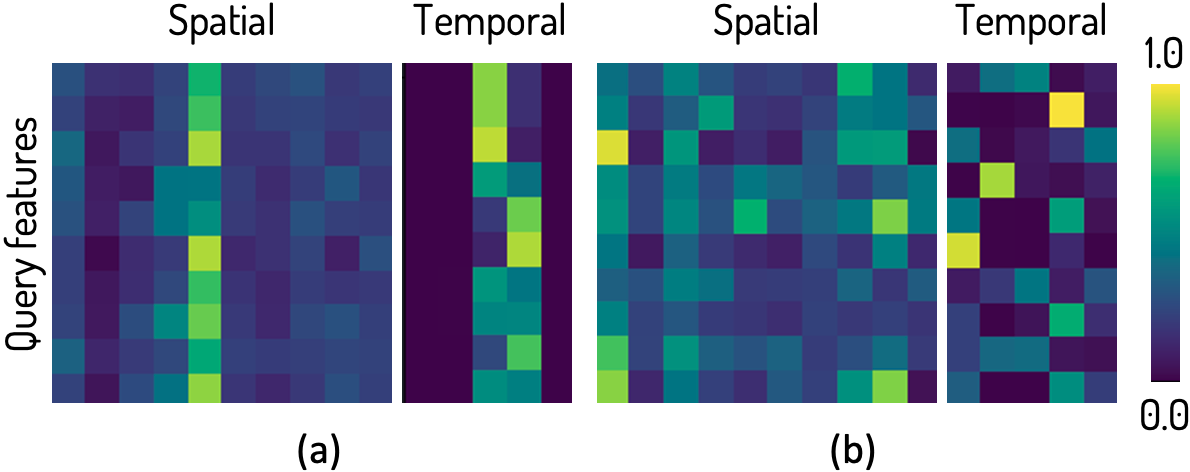}	
			\caption{Matching probability maps of spatial and temporal memories, when they are trained (a) without and (b) with the memory spread loss. We randomly select 10 query features from a gallery set of MARS~\cite{zheng2016mars}. We can see that the memory spread loss encourages our model to access all items in the memories. (Best viewed in color.)}
			\label{fig:mem_spread}
			\vspace{-0.2cm}
		\end{figure}

	\subsection{Discussion} \label{subsec:discussion}
	\vspace{-0.1cm}
	
		\paragraph{Ablation study.}

			We show in Table~\ref{table:ablation_studies} an ablation study of our model on MARS~\cite{zheng2016mars}, DukeV~\cite{wu2018exploit}, and LS-VID~\cite{li2019global} in terms of rank-$1$ accuracy(\%) and mAP(\%). For the baseline, we use the same network architecture as the encoder, while removing two heads for query maps, and exploit global and temporal average pooling to aggregate person representations. From \textcircled{\raisebox{-0.9pt}{1}} and \textcircled{\raisebox{-0.9pt}{3}}, we can clearly see that the feature refinement process using a spatial memory boosts the reID performance, while \textcircled{\raisebox{-0.9pt}{1}} and \textcircled{\raisebox{-0.9pt}{5}} demonstrate that using a temporal memory for aggregating frame-level representations gives better results. \textcircled{\raisebox{-0.9pt}{3}}, \textcircled{\raisebox{-0.9pt}{5}}, and \textcircled{\raisebox{-0.9pt}{7}} further show the spatial and temporal memories are complementary to each other. Note that LS-VID provides person videos with more diverse spatial and temporal distractors than the other datasets. It contains videos of three times larger number of identities than MARS and DukeV, which are captured under two times larger number of cameras. Our memories help the baseline model to handle such distractors, giving the significant performance gains on LS-VID. The performance gains by the memories are relatively small on DukeV, since it contains person videos that are manually annotated by humans,~\ie,~with less distractors, where the simple baseline already gives 95\% rank-1 accuracy. By comparing \textcircled{\raisebox{-0.9pt}{2}} to \textcircled{\raisebox{-0.9pt}{3}}, \textcircled{\raisebox{-0.9pt}{4}} to \textcircled{\raisebox{-0.9pt}{5}}, and \textcircled{\raisebox{-0.9pt}{6}} to \textcircled{\raisebox{-0.9pt}{7}}, we can see that enforcing our model to address all memory items during training by a memory spread loss consistently enhances the performance.
						
			To further verify the effectiveness of the memory spread loss, we visualize matching probability maps of spatial and temporal memories on MARS, when the memories are trained without~(Fig.~\ref{fig:mem_spread}(a)) and with~(Fig.~\ref{fig:mem_spread}(b)) the loss. We randomly choose frames or sequences from a gallery set of MARS, and extract query features, $\mathbf{q}^{\text s}_{i,k}$ and $\mathbf{q}^{\text t}$, from them. We then compute matching probabilities with keys of the spatial and temporal memories using Eq.~\eqref{eq:a_s} and Eq.~\eqref{eq:a_t}, respectively. We can see that the memory spread loss encourages our model to leverage all items in the memories, while preventing it from accessing particular items only. This enables our spatial and temporal memories to produce diverse outputs depending on both frame-level scene details and sequence-level temporal contexts.
			
		\vspace{-0.5cm}
		\paragraph{Spatial memory.}

		\begin{figure}[t]
			\begin{minipage}{0.25\textwidth}
				\centering
				\includegraphics[height=0.85\linewidth,width=0.99\linewidth]{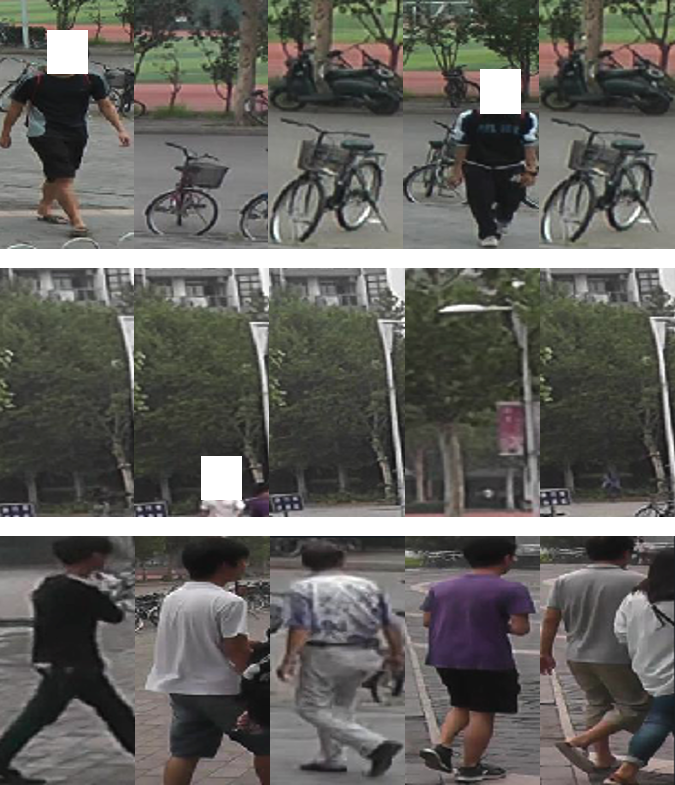}
				\caption{Top-$5$ retrieved frames from a gallery set of MARS~\cite{zheng2016mars}, whose query features have high matching probabilities with a key of the spatial memory.}
			  	\label{fig:smem_key}
			\end{minipage}
			\begin{minipage}{0.22\textwidth}
				\centering
				\includegraphics[height=0.97\linewidth,width=0.99\linewidth]{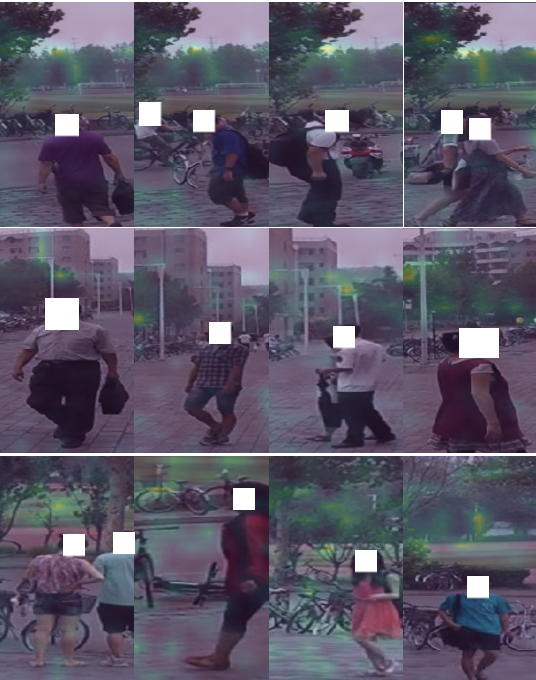}
				\caption{The magnitude difference of person representations before and after the refinement using the spatial memory.}
			  	\label{fig:smem_attention}
			\end{minipage}\hfill
		   \vspace{-0.2cm}
		\end{figure}

\begin{figure}[t]
	    	\centering
	    	\includegraphics[width=0.95\linewidth]{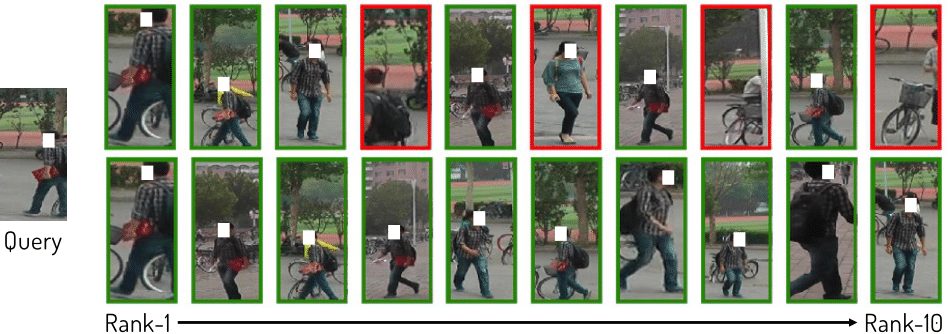}
	    	\caption{Comparison of top-$10$ retrieval results on the test split of MARS~\cite{zheng2016mars} using the original frame-level features $\mathbf{f}^{\text o}_i$ (top) and refined ones $\mathbf{f}^{\text s}_i$ (bottom). Results with green boxes have the same identity as the query, while those with red boxes do not. We show the first frame of sequences for the purpose of visualization. (Best viewed in color.)}
	    	\label{fig:smem_val}
	    	\vspace{-0.5cm}
	    \end{figure}

			In Fig.~\ref{fig:smem_key}, we visualize video frames whose query features $\mathbf{q}^{\text s}_{i,k}$ have high matching probabilities with randomly chosen keys from the spatial memory~(see Eq.~\eqref{eq:a_s}). We can observe that each key retrieves the video frames that share similar scene details such as a play field~(1st row), a street light~(2nd row), or concrete pavers~(3rd row). This verifies that our model accesses the spatial memory depending on the scene details for each video frame. The spatial memory aggregates the features for scene details, and we use them to refine frame-level person representations~(see Eq.~\eqref{eq:refine}). To see the effect of the refinement, we visualize in Fig.~\ref{fig:smem_attention} the magnitude difference of person representations, overlaid on input images, using bilinear interpolation, before and after the refinement,~\ie, $\big\{\lVert \mathbf{f}^\text{s}_{i,k} \rVert_2 - \lVert \mathbf{f}^\text{o}_{i,k} \rVert_2 \big| k \in H \times W \big\}$. We can observe that the differences mainly occur from distracting scene details,~\eg, concrete pavers, playfield, or street lights, implying that the memory suppresses features from them. Note that the video frames in the 1st row of Fig.~\ref{fig:smem_attention} share the same background while pedestrians appear in different positions. However, regardless of the person's position, the memory removes features from background clutter. Figure~\ref{fig:smem_val} compares retrieval results when we use initial person representations $\mathbf{f}^{\text o}_i$ (top) and refined ones $\mathbf{f}^{\text s}_i$ (bottom). Note that we use global and temporal average pooling to obtain the person representations, instead of exploiting the temporal memory, to see the effect of the refinement by the spatial memory. We can see that the initial representations retrieve person sequences of different identities from the query but with similar scene details~(\eg,~a play field). On the other hand, the refined ones retrieve person sequences with the same identity as the query correctly, regardless of background clutter in each frame. This also suggests that the refinement process using the spatial memory suppresses information of the scene details in person representations.

		\begin{figure}[t!]
			\centering
			\renewcommand*{\thesubfigure}{}
			\hspace{-0.3cm}
			\subfigure[]{
				\begin{minipage}[t]{0.46\linewidth}
					\centering
					\includegraphics[height=0.9\linewidth,width=\linewidth]{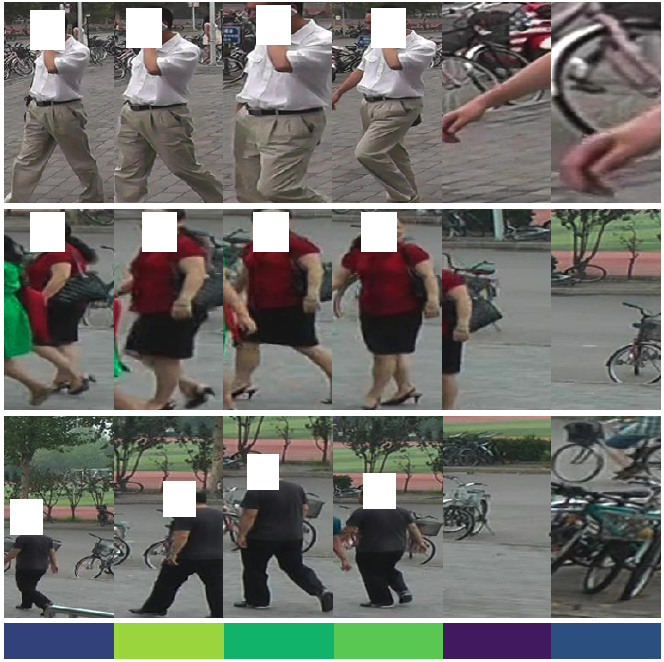}
				\end{minipage}
			}
			\hspace{-0.3cm}
			\subfigure[]{
				\begin{minipage}[t]{0.46\linewidth}
					\centering
					\includegraphics[height=0.9\linewidth,width=\linewidth]{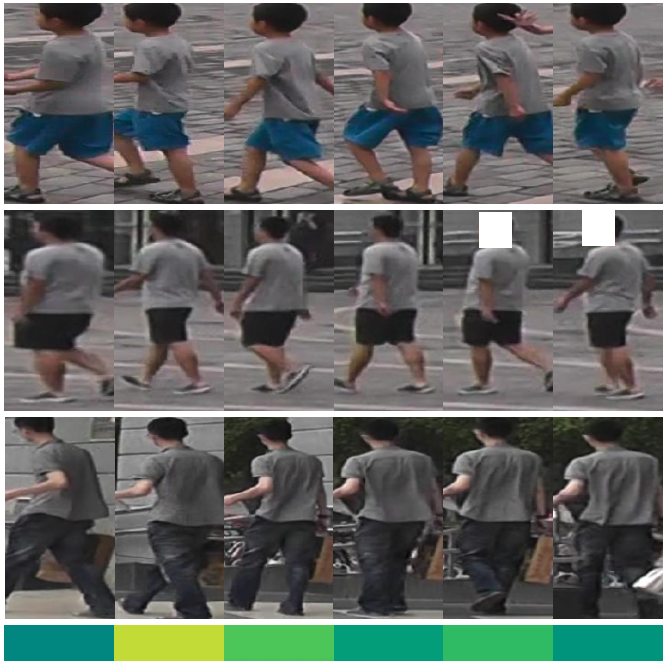}
				\end{minipage}
			}
			\hspace{-0.2cm}
			\subfigure[]{
				\begin{minipage}[t]{0.027\linewidth}
					\centering
					\includegraphics[height=15\linewidth,width=\linewidth]{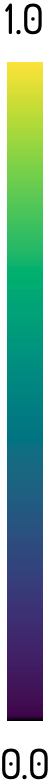}
				\end{minipage}
			}
			\vspace{-0.5cm}
			\caption{Examples of sequences from a gallery set of MARS~\cite{zheng2016mars}, whose query features show high matching probabilities with a particular key in the temporal memory. We also visualize temporal attentions stored in corresponding values of the memory. (Best viewed in color.)}
			\label{fig:tmem_key}
			\vspace{-0.6cm}
		\end{figure}
		
		\begin{figure}
			\centering
			\vspace{-0.2cm}
			\renewcommand*{\thesubfigure}{}
			\hspace{-0.3cm}
			\subfigure[]{
				\begin{minipage}[t]{0.95\linewidth}
					\centering
					\includegraphics[height=0.6\linewidth,width=\linewidth]{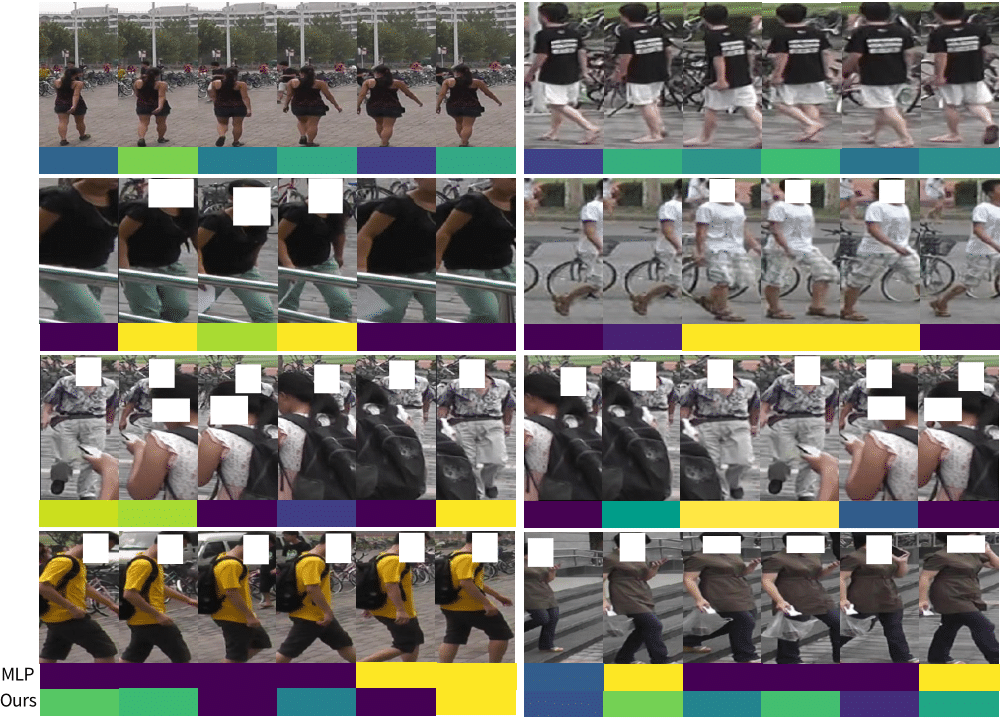}
				\end{minipage}
			}
			\hspace{-0.25cm}
			\subfigure[]{
				\begin{minipage}[t]{0.04\linewidth}
					\centering
					\includegraphics[height=13.6\linewidth,width=\linewidth]{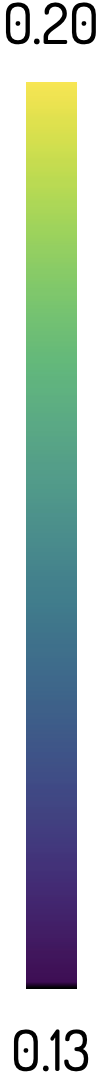}
				\end{minipage}
			}
			\vspace{-0.6cm}
			\caption{Examples of temporal attentions generated by the temporal memory on the test split of MARS~\cite{zheng2016mars}. Note that the sequence on the right side of the 3rd row is made by reordering the sequence on the left side. (Best viewed in color.)}
			\label{fig:tmem_val}
			\vspace{-0.6cm}
		\end{figure}

		\vspace{-0.5cm}
		\paragraph{Temporal memory.}
		
			We visualize in Fig.~\ref{fig:tmem_key} person sequences whose query features $\mathbf{q}^{\text t}$ show high matching probabilities (see Eq.~\eqref{eq:a_t}) with randomly chosen keys from the temporal memory. We also visualize the corresponding values of the memory in the below. We can observe that each key retrieves the sequences with similar temporal patterns,~\eg,~persons disappear at the end of the sequence~(left) or appear in all frames with similar appearances~(right), and the values highlight discriminative frames in each sequence. This verifies that the keys encode prototypes of temporal patterns in person videos, and the values of the memory store temporal attentions which are optimized for the corresponding temporal patterns. Note that we aggregate individual values of the memory with matching probabilities between keys and input query features to synthesize temporal attentions specific for input person sequences~(see Eq.~\eqref{eq:att}). Figure~\ref{fig:tmem_val} shows examples of the aggregated temporal attentions. When the temporal memory takes a sequence with less temporal distractors as an input, the memory generates similar attentions for all frames~(1st row). Namely, the memory works similarly to the temporal average pooling which fuses video frames with equal probabilities. On the other hand, in case of a sequence with severe temporal distractors,~\eg,~misalignments between frames~(2nd row) or occlusions~(3rd row), the memory lowers attentions for the frames where such variations occur, suggesting that the temporal memory allows our model to extract person representations robust to the temporal variations. Note that we can replace the temporal memory with multilayer perceptrons (MLPs) by directly regressing attentions from the encoded context $\mathbf{q}^{\text t}$~\cite{li2018diversity}. To compare this approach with ours, we use two-layer perceptrons whose sizes are $2048 \times N$ and $N \times L$, respectively, which makes the number of parameters the same as ours. We found that MLP often produce attentions that focus more on few certain frames, ignoring features from the other frames (see the last row of Fig.~\ref{fig:tmem_val}), and this leads to large performance drops, 1.3/1.5 (R-1/mAP) on MARS. The result are similar, even the size of MLPs increase~(\eg,~2048x512 and 512x6). These show the effectiveness of our approach that predicts attentions by discovering repetitive temporal patterns in a dataset and searching the most relevant patterns to the context of an input video.

	\subsection{Comparison with the state of the art} \label{subsec:comp_w_sota}
	\vspace{-0.1cm}
	
		We compare in Table~\ref{table:sota} STMN with the state of the art in terms of rank-1 accuracy and mAP on MARS~\cite{zheng2016mars}, DukeV~\cite{wu2018exploit} and LS-VID~\cite{li2019global}. We found that previous methods compare their performance using different test strategies. For fair comparisons, we classify them into two groups, depending on whether they follow RRS or all-frames strategies for evaluation. The methods,~\eg,~\cite{li2018diversity,liu2019spatially}, which follow the RRS strategy~\cite{li2018diversity}, divide an input video into $L$ chunks of equal length. They then sample the first frame of each chunk to obtain a sequence of $L$ frames, regardless of the total number of frames. On the other hand, several works use all frames in an input video by grouping them into multiple sequences of length $L$. They extract a person representation from each sequence independently, and average all the representations to represent the input video. Note that we reproduce TCLNet~\cite{hou2020temporal} and  MGH~\cite{yan2020learning} to evaluate them on the both strategies. Using all frames in given videos to extract person representations does give performance gains for TCLNet, MGH, and STMN. This, however, is far from practical usages in that it runs,~\eg,~35 times slower than the RRS strategy on LS-VID, requiring more than three hours for evaluation using a Titan RTX 2080Ti GPU. Furthermore, the time for searching persons increases linearly as the number of video frames increases.
		
		From Table~\ref{table:sota}, we have following observations: 1) On the RRS setting, STMN sets a new state of the art on the three benchmarks. The results of STMN using the RRS even surpass those of previous methods,~\eg,~COSAM~\cite{subramaniam2019co}, M3D~\cite{li2019multi}, and GLTR~\cite{li2019global} on the all-frames setting. This suggests that STMN already extracts essential information for identifying a person with sampled frames only, showing its efficiency over the previous methods. This characteristic is crucial for massive surveillance systems which need to search for a person of interest from lots of videos in a very short time; 2) DRSA~\cite{li2018diversity} leverages attention modules for handling spatial and temporal distractors in videos, while STMN exploits spatial and temporal memories instead. The performance gap between these two methods demonstrates the superiority of our framework over the attention-based method; 3) Co-attention-based methods~\cite{liu2019spatially,li2019global,yan2020learning} may propagate non-discriminative features across frames when multiple frames share common background clutter or occlusion. As a result, there are large performance gaps between these methods and STMN on LS-VID, the most challenging dataset, which contains sequences captured under various conditions~(\emph{e.g.}, lighting/background changes, indoor/outdoor changes) with frequent occlusions; 4) TCLNet~\cite{hou2020temporal} and MGH~\cite{yan2020learning} are the most recently introduced video reID methods. They boost the reID performance using a temporal saliency erasing module and a multi-granular hypergraph, respectively. They, however, give results worse than STMN on the RRS setting. By using all frames, they may show comparable results to STMN, however note that the size of a person representation is much larger than that of STMN~(TCLNet:$4,096$, MGH:$5,120$ vs. STMN:$2,048$).
		
		\begin{table}[!t]
			\centering
			\huge
			\resizebox{0.98\linewidth}{!}{
				\begin{tabular}{clcccccc}
				\hline
				& \multicolumn{1}{c}{\multirow{2}*{Methods}} & \multicolumn{2}{c}{MARS} & \multicolumn{2}{c}{DukeV} & \multicolumn{2}{c}{LS-VID} \\
				& & rank-1 & mAP & rank-1 & mAP & rank-1 & mAP\\
				\hline
				\parbox[t]{10mm}{\multirow{10}{*}{\rotatebox[origin=c]{90}{RRS}}}
				& EUG~\cite{wu2018exploit}			& 62.7 & 42.5 & 72.8 & 63.2 & -    & -    \\
				& SeeForest~\cite{zhou2017see}		& 70.6 & 50.7 & -    & -    & -    & -    \\
				& QAN~\cite{liu2017quality}			& 73.7 & 51.7 & -    & -    & -    & -    \\
				& DRSA~\cite{li2018diversity}		& 82.3 & 65.8 & -    & -    & -    & -    \\
				& CSA~\cite{chen2018video} 			& 86.3 & 76.1 & -    & -    & -    & -    \\
				& STE-NVAN~\cite{liu2019spatially}	& 88.9 & 81.2 & 95.2 & \underline{93.5} & (72.1) & (56.6) \\
				& TCLNet~\cite{hou2020temporal}		& (88.5) & (80.9) & (95.0) & (92.8) & (75.0) & (\underline{60.2}) \\
				& MGH~\cite{yan2020learning}			& (\underline{89.2}) & (\underline{83.4}) & (\underline{95.3}) & (93.4) & (\underline{75.3}) & (58.9) \\
				& STMN 								& \textbf{89.9} & \textbf{83.7} & \textbf{96.7} & \textbf{94.6} & \textbf{80.6} & \textbf{66.6} \\
				\hline
				\parbox[t]{10mm}{\multirow{8}{*}{\rotatebox[origin=c]{90}{All frames}}}
				& COSAM~\cite{subramaniam2019co}		& 83.7 & 77.2 & 94.4 & 94.0 & -    & -    \\
				& STMP~\cite{liu2019spatial}	 		& 84.4 & 72.7 & -    & -    & 56.8 & 39.1 \\
				& M3D~\cite{li2019multi} 			& 84.4 & 74.1 & -    & -    & 57.7 & 40.1 \\
				& Part-Aligned~\cite{suh2018part}	& 84.7 & 75.9 & -    & -    & -    & -    \\
				& STA~\cite{fu2019sta} 				& 86.3 & 80.8 & 96.0 & 95.0 & -    & -   \\
				& GLTR~\cite{li2019global} 			& 87.0 & 78.5 & 96.3 & 93.7 & 63.1 & 44.3 \\
				& TCLNet~\cite{hou2020temporal}		& (89.1) & (83.4) & (\underline{96.7}) & (\underline{95.6}) & (\underline{81.0}) & (\underline{67.2}) \\
				& MGH~\cite{yan2020learning}			& (\underline{89.4}) & (\textbf{85.3}) & (95.0) & (94.6) & (79.6) & (61.8) \\
				& STMN 								& \textbf{90.5} & \underline{84.5} & \textbf{97.0} & \textbf{95.9} & \textbf{82.1} & \textbf{69.2} \\
				\hline
				\end{tabular}
			}
			\vspace{+0.1cm}
			\caption{Comparison with the state of the art on MARS~\cite{zheng2016mars}, DukeV~\cite{wu2018exploit}, and LS-VID~\cite{li2019global} in terms of rank-$1$ accuracy(\%) and mAP(\%). Numbers in bold indicate the best performance and underscored ones are the second best. Results in brackets are obtained with the source codes provided by the authors.}
			\label{table:sota}
			\vspace{-0.4cm}
		\end{table}
								
\section{Conclusion}
\vspace{-0.1cm}
We have presented a novel video-based person reID method, dubbed STMN, that extracts robust person representations against spatial and temporal distractors in videos. To this end, we have proposed to exploit two external memory networks, spatial and temporal memories, to refine frame-level representations and to aggregate them into a sequence one, focusing on discriminative frames. We have also proposed a memory spread loss that prevents certain memory items from remaining redundant. We have shown that STMN achieves state-of-the-art performance on standard video-based reID benchmarks, and demonstrated the effectiveness of each component of our method with an extensive ablation study.

\paragraph{Acknowledgments}
This research was supported in part by the National Research Foundation of Korea (NRF) grant funded by the Korea government (MSIP) (NRF-2019R1A2C2084816) and the Yonsei University Research Fund of 2021 (2021-22-0001).
\newpage

{\small
\bibliographystyle{ieee_fullname}
\bibliography{egbib}

\begin{thebibliography}{10}\itemsep=-1pt

\bibitem{cai2018memory}
Qi Cai, Yingwei Pan, Ting Yao, Chenggang Yan, and Tao Mei.
\newblock Memory matching networks for one-shot image recognition.
\newblock In {\em CVPR}, 2018.

\bibitem{chen2018video}
Dapeng Chen, Hongsheng Li, Tong Xiao, Shuai Yi, and Xiaogang Wang.
\newblock Video person re-identification with competitive snippet-similarity
  aggregation and co-attentive snippet embedding.
\newblock In {\em CVPR}, 2018.

\bibitem{chentemporal}
Guangyi Chen, Yongming Rao, Jiwen Lu, and Jie Zhou.
\newblock Temporal coherence or temporal motion: Which is more critical for
  video-based person re-identification?
\newblock In {\em ECCV}, 2020.

\bibitem{eom2019learning}
Chanho Eom and Bumsub Ham.
\newblock Learning disentangled representation for robust person
  re-identification.
\newblock In {\em NeurIPS}, 2019.

\bibitem{fu2019sta}
Yang Fu, Xiaoyang Wang, Yunchao Wei, and Thomas Huang.
\newblock S{TA}: Spatial-temporal attention for large-scale video-based person
  re-identification.
\newblock In {\em AAAI}, 2019.

\bibitem{gong2019memorizing}
Dong Gong, Lingqiao Liu, Vuong Le, Budhaditya Saha, Moussa~Reda Mansour, Svetha
  Venkatesh, and Anton van~den Hengel.
\newblock Memorizing normality to detect anomaly: Memory-augmented deep
  autoencoder for unsupervised anomaly detection.
\newblock In {\em ICCV}, 2019.

\bibitem{he2016deep}
Kaiming He, Xiangyu Zhang, Shaoqing Ren, and Jian Sun.
\newblock Deep residual learning for image recognition.
\newblock In {\em CVPR}, 2016.

\bibitem{hermans2017defense}
Alexander Hermans, Lucas Beyer, and Bastian Leibe.
\newblock In defense of the triplet loss for person re-identification.
\newblock {\em arXiv:1703.07737}, 2017.

\bibitem{hirzer2011person}
Martin Hirzer, Csaba Beleznai, Peter~M Roth, and Horst Bischof.
\newblock Person re-identification by descriptive and discriminative
  classification.
\newblock In {\em SCIA}, 2011.

\bibitem{hochreiter1997long}
Sepp Hochreiter and J{\"u}rgen Schmidhuber.
\newblock Long short-term memory.
\newblock {\em Neural computation}.

\bibitem{hou2020temporal}
Ruibing Hou, Hong Chang, Bingpeng Ma, Shiguang Shan, and Xilin Chen.
\newblock Temporal complementary learning for video person re-identification.
\newblock In {\em ECCV}, 2020.

\bibitem{kingma2014adam}
Diederik~P Kingma and Jimmy Ba.
\newblock Adam: A method for stochastic optimization.
\newblock In {\em ICLR}, 2015.

\bibitem{kipf2017semi}
Thomas~N Kipf and Max Welling.
\newblock Semi-supervised classification with graph convolutional networks.
\newblock In {\em ICLR}, 2017.

\bibitem{krizhevsky2012imagenet}
Alex Krizhevsky, Ilya Sutskever, and Geoffrey~E Hinton.
\newblock Image{N}et classification with deep convolutional neural networks.
\newblock In {\em NIPS}, 2012.

\bibitem{lee2015deeply}
Chen-Yu Lee, Saining Xie, Patrick Gallagher, Zhengyou Zhang, and Zhuowen Tu.
\newblock Deeply-supervised nets.
\newblock In {\em Artificial {I}ntelligence and {S}tatistics}, 2015.

\bibitem{li2019global}
Jianing Li, Jingdong Wang, Qi Tian, Wen Gao, and Shiliang Zhang.
\newblock Global-local temporal representations for video person
  re-identification.
\newblock In {\em ICCV}, 2019.

\bibitem{li2019multi}
Jianing Li, Shiliang Zhang, and Tiejun Huang.
\newblock Multi-scale 3{D} convolution network for video based person
  re-identification.
\newblock In {\em AAAI}, 2019.

\bibitem{li2018diversity}
Shuang Li, Slawomir Bak, Peter Carr, and Xiaogang Wang.
\newblock Diversity regularized spatiotemporal attention for video-based person
  re-identification.
\newblock In {\em CVPR}, 2018.

\bibitem{li2018harmonious}
Wei Li, Xiatian Zhu, and Shaogang Gong.
\newblock Harmonious attention network for person re-identification.
\newblock In {\em CVPR}, 2018.

\bibitem{liu2019spatially}
Chih-Ting Liu, Chih-Wei Wu, Yu-Chiang~Frank Wang, and Shao-Yi Chien.
\newblock Spatially and temporally efficient non-local attention network for
  video-based person re-identification.
\newblock In {\em BMVC}, 2019.

\bibitem{liu2017hydraplus}
Xihui Liu, Haiyu Zhao, Maoqing Tian, Lu Sheng, Jing Shao, Shuai Yi, Junjie Yan,
  and Xiaogang Wang.
\newblock Hydraplus-net: Attentive deep features for pedestrian analysis.
\newblock In {\em ICCV}, 2017.

\bibitem{liu2017quality}
Yu Liu, Junjie Yan, and Wanli Ouyang.
\newblock Quality aware network for set to set recognition.
\newblock In {\em CVPR}, 2017.

\bibitem{liu2019spatial}
Yiheng Liu, Zhenxun Yuan, Wengang Zhou, and Houqiang Li.
\newblock Spatial and temporal mutual promotion for video-based person
  re-identification.
\newblock In {\em AAAI}, 2019.

\bibitem{mclaughlin2016recurrent}
Niall McLaughlin, Jesus Martinez~del Rincon, and Paul Miller.
\newblock Recurrent convolutional network for video-based person
  re-identification.
\newblock In {\em CVPR}, 2016.

\bibitem{miller2016key}
Alexander Miller, Adam Fisch, Jesse Dodge, Amir-Hossein Karimi, Antoine Bordes,
  and Jason Weston.
\newblock Key-value memory networks for directly reading documents.
\newblock In {\em EMNLP}, 2016.

\bibitem{oh2019video}
Seoung~Wug Oh, Joon-Young Lee, Ning Xu, and Seon~Joo Kim.
\newblock Video object segmentation using space-time memory networks.
\newblock In {\em ICCV}, 2019.

\bibitem{park2020learning}
Hyunjong Park, Jongyoun Noh, and Bumsub Ham.
\newblock Learning memory-guided normality for anomaly detection.
\newblock In {\em CVPR}, 2020.

\bibitem{ristani2016performance}
Ergys Ristani, Francesco Solera, Roger Zou, Rita Cucchiara, and Carlo Tomasi.
\newblock Performance measures and a data set for multi-target, multi-camera
  tracking.
\newblock In {\em ECCV}, 2016.

\bibitem{su2017pose}
Chi Su, Jianing Li, Shiliang Zhang, Junliang Xing, Wen Gao, and Qi Tian.
\newblock Pose-driven deep convolutional model for person re-identification.
\newblock In {\em ICCV}, 2017.

\bibitem{subramaniam2019co}
Arulkumar Subramaniam, Athira Nambiar, and Anurag Mittal.
\newblock Co-segmentation inspired attention networks for video-based person
  re-identification.
\newblock In {\em ICCV}, 2019.

\bibitem{suh2018part}
Yumin Suh, Jingdong Wang, Siyu Tang, Tao Mei, and Kyoung Mu~Lee.
\newblock Part-aligned bilinear representations for person re-identification.
\newblock In {\em ECCV}, 2018.

\bibitem{sukhbaatar2015end}
Sainbayar Sukhbaatar, Jason Weston, Rob Fergus, et~al.
\newblock End-to-end memory networks.
\newblock In {\em NeurIPS}, 2015.

\bibitem{vaswani2017attention}
Ashish Vaswani, Noam Shazeer, Niki Parmar, Jakob Uszkoreit, Llion Jones,
  Aidan~N Gomez, {\L}ukasz Kaiser, and Illia Polosukhin.
\newblock Attention is all you need.
\newblock In {\em NeurIPS}, 2017.

\bibitem{wang2015training}
Liwei Wang, Chen-Yu Lee, Zhuowen Tu, and Svetlana Lazebnik.
\newblock Training deeper convolutional networks with deep supervision.
\newblock {\em arXiv:1505.02496}, 2015.

\bibitem{wang2014person}
Taiqing Wang, Shaogang Gong, Xiatian Zhu, and Shengjin Wang.
\newblock Person re-identification by video ranking.
\newblock In {\em ECCV}, 2014.

\bibitem{wang2018non}
Xiaolong Wang, Ross Girshick, Abhinav Gupta, and Kaiming He.
\newblock Non-local neural networks.
\newblock In {\em CVPR}, 2018.

\bibitem{weston2014memory}
Jason Weston, Sumit Chopra, and Antoine Bordes.
\newblock Memory networks.
\newblock In {\em ICLR}, 2015.

\bibitem{wu2018exploit}
Yu Wu, Yutian Lin, Xuanyi Dong, Yan Yan, Wanli Ouyang, and Yi Yang.
\newblock Exploit the unknown gradually: One-shot video-based person
  re-identification by stepwise learning.
\newblock In {\em CVPR}, 2018.

\bibitem{yan2016person}
Yichao Yan, Bingbing Ni, Zhichao Song, Chao Ma, Yan Yan, and Xiaokang Yang.
\newblock Person re-identification via recurrent feature aggregation.
\newblock In {\em ECCV}, 2016.

\bibitem{yan2020learning}
Yichao Yan, Jie Qin, Jiaxin Chen, Li Liu, Fan Zhu, Ying Tai, and Ling Shao.
\newblock Learning multi-granular hypergraphs for video-based person
  re-identification.
\newblock In {\em CVPR}, 2020.

\bibitem{yang2020spatial}
Jinrui Yang, Wei-Shi Zheng, Qize Yang, Ying-Cong Chen, and Qi Tian.
\newblock Spatial-temporal graph convolutional network for video-based person
  re-identification.
\newblock In {\em CVPR}, 2020.

\bibitem{yoo2019coloring}
Seungjoo Yoo, Hyojin Bahng, Sunghyo Chung, Junsoo Lee, Jaehyuk Chang, and
  Jaegul Choo.
\newblock Coloring with limited data: Few-shot colorization via memory
  augmented networks.
\newblock In {\em CVPR}, 2019.

\bibitem{zhang2020multi}
Zhizheng Zhang, Cuiling Lan, Wenjun Zeng, and Zhibo Chen.
\newblock Multi-granularity reference-aided attentive feature aggregation for
  video-based person re-identification.
\newblock In {\em CVPR}, 2020.

\bibitem{zhao2017spindle}
Haiyu Zhao, Maoqing Tian, Shuyang Sun, Jing Shao, Junjie Yan, Shuai Yi,
  Xiaogang Wang, and Xiaoou Tang.
\newblock Spindle{N}et: Person re-identification with human body region guided
  feature decomposition and fusion.
\newblock In {\em CVPR}, 2017.

\bibitem{zheng2016mars}
Liang Zheng, Zhi Bie, Yifan Sun, Jingdong Wang, Chi Su, Shengjin Wang, and Qi
  Tian.
\newblock M{ARS}: A video benchmark for large-scale person re-identification.
\newblock In {\em ECCV}.

\bibitem{zheng2019joint}
Zhedong Zheng, Xiaodong Yang, Zhiding Yu, Liang Zheng, Yi Yang, and Jan Kautz.
\newblock Joint discriminative and generative learning for person
  re-identification.
\newblock In {\em CVPR}, 2019.

\bibitem{zhong2017random}
Zhun Zhong, Liang Zheng, Guoliang Kang, Shaozi Li, and Yi Yang.
\newblock Random erasing data augmentation.
\newblock {\em arXiv:1708.04896}, 2017.

\bibitem{zhong2019invariance}
Zhun Zhong, Liang Zheng, Zhiming Luo, Shaozi Li, and Yi Yang.
\newblock Invariance matters: Exemplar memory for domain adaptive person
  re-identification.
\newblock In {\em CVPR}, 2019.

\bibitem{zhou2017see}
Zhen Zhou, Yan Huang, Wei Wang, Liang Wang, and Tieniu Tan.
\newblock See the forest for the trees: Joint spatial and temporal recurrent
  neural networks for video-based person re-identification.
\newblock In {\em CVPR}, 2017.

\end{thebibliography}
}

\end{document}